\documentclass[journal]{IEEEtran}

\ifCLASSINFOpdf
  
\else
 
\fi

\usepackage{amsmath}
\usepackage{graphicx, subfig, color, multirow, amsmath}
\graphicspath{{figs/}}
\usepackage{caption}
\usepackage{bbding}
\usepackage{multirow}
\usepackage{float}
\usepackage{booktabs}
\usepackage[edges]{forest}
\usetikzlibrary{shadows}
\begin{document}

\title{Spatiotemporal Attention Learning Framework for Event-Driven Object Recognition}

\author{Tiantian~Xie, Pengpai~Wang and~Rosa~H.~M.~Chan,~\IEEEmembership{Senior Member},~IEEE

\thanks{T. Xie and R. H. M. Chan are with the Department of Electrical Engineering, City University of Hong Kong, e-mail: rosachan@cityu.edu.hk. Pengpai Wang is with the Shenzhen Research Institute, City University of Hong Kong, Shenzhen, China. Pengpai Wang and Rosa H.M. Chan are with the State Key Laboratory of Terahertz and Millimeter Waves, City University of Hong Kong, Hong Kong, China.}
}

%
%

\markboth{Journal of \LaTeX\ Class Files,~Vol.~14, No.~8, August~2024}%
{Shell \MakeLowercase{\textit{et al.}}: Bare Demo of IEEEtran.cls for IEEE Journals}

\maketitle

\begin{abstract}

Event-based vision sensors, inspired by biological neural systems, asynchronously capture local pixel-level intensity changes as a sparse event stream containing position, polarity, and timestamp information. These neuromorphic sensors offer significant advantages in dynamic range, latency, and power efficiency. Their working principle inherently addresses traditional camera limitations such as motion blur and redundant background information, making them particularly suitable for dynamic vision tasks. While recent works have proposed increasingly complex event-based architectures, the computational overhead and parameter complexity of these approaches limit their practical deployment. This paper presents a novel spatiotemporal learning framework for event-based object recognition, utilizing a VGG network enhanced with Convolutional Block Attention Module (CBAM). Our approach achieves comparable performance to state-of-the-art ResNet-based methods while reducing parameter count by 2.3\% compared to the original VGG model. Specifically, it outperforms ResNet-based methods like MVF-Net, achieving the highest Top-1 accuracy of 76.4\% (pretrained) and 71.3\% (not pretrained) on CIFAR10-DVS, and 72.4\% (not pretrained) on N-Caltech101. These results highlight the robustness of our method when pretrained weights are not used, making it suitable for scenarios where transfer learning is unavailable. Moreover, our approach reduces reliance on data augmentation. Experimental results on standard event-based datasets demonstrate the framework's efficiency and effectiveness for real-world applications.

\end{abstract}


\IEEEpeerreviewmaketitle

\section{INTRODUCTION}

\IEEEPARstart Object recognition has always been a significant research area within the field of computer vision due to its extensive applications in medical diagnostics, industry manufacturing, human-computer interaction domain. With the great development of the deep learning recently, feature extraction and recognition algorithms have already made remarkable achievements. The event camera is proposed as a neuromorphic vision sensor that designed to trigger "event" asynchronously in terms of whether the brightness change per-pixel exceeds the threshold \cite{1,2,3}. The output event is single point that can be represented as a quadruple $e_i=\{x, y, t, p\}$ containing the spatial position, timestamp and polarity information. Obviously, this mechanism enables the event camera to solely concentrate on the dynamic outline while discarding the redundant and static background. Its advantages include high time resolution (microsecond-level), high dynamic range, low power consumption, low motion blur, etc, and has been widely utilized in the light flow estimation, attitude estimation, 3D reconstruction\cite{ercan2023evreal}, automatic driving\cite{shariff2024event} and other fields. 

However, the single event point typically lacks the meaningful representation ability, and the event stream is essentially a set of discrete points possessing correlation in the spatial-temporal domain. Consequently, extracting the spatial-temporal correspondence of the discrete events plays a crucial role in this task. To accommodate the existed mature vision algorithms, most work resort to adopt the image-like representations. Among them, the commonly employed methods contain the event-frame representations \cite{5,6,7}, time surface \cite{delbruck2008frame}, or the fusion representations\cite{el2023cstr}. While in the realm of neural network, variety of complicated mechanism or module have been integrated into the end-to-end learning framework \cite{p2, p1}.  However, merely expanding the depth or complexity of the network structure can obtain satisfactory performance to a certain extent, but theirs huge number of parameters and complicated implementation of algorithm have rendered a certain demand for the computing resources. Besides, the hardness of training the spiking neural network and its intrinsic high requirements on hardware suitability have been an hinder for the development.

In consequence, blindly proposing more complex representations or network structures is not the optimal solution to the problem. In this paper, we aim to establish a concise overall architecture for the event-based recognition. Consequently, we equipped the VGG network
with the convolutional block attention (CBAM) module \cite{simonyan2014very, woo2018cbam}, which is enable to effectively extract the spatial-temporal feature while maintaining a simple network structure. We adopted the SpikingJelly open-source framework for event data pre-processing \cite{fang2023spikingjelly} , whose parameters can be adjusted flexibly and expansively. Our proposed framework can not only achieves satisfactory accuracy on CIFAR10-DVS\cite{li2017cifar10}and N-Caltech101 dataset\cite{orchard2015converting}, but also provides inspiration for developing more accurate and energy-saving  event-based recognition systems in the future.

\begin{figure*}[htbp]
	\centering
    \includegraphics[width=0.9\textwidth]{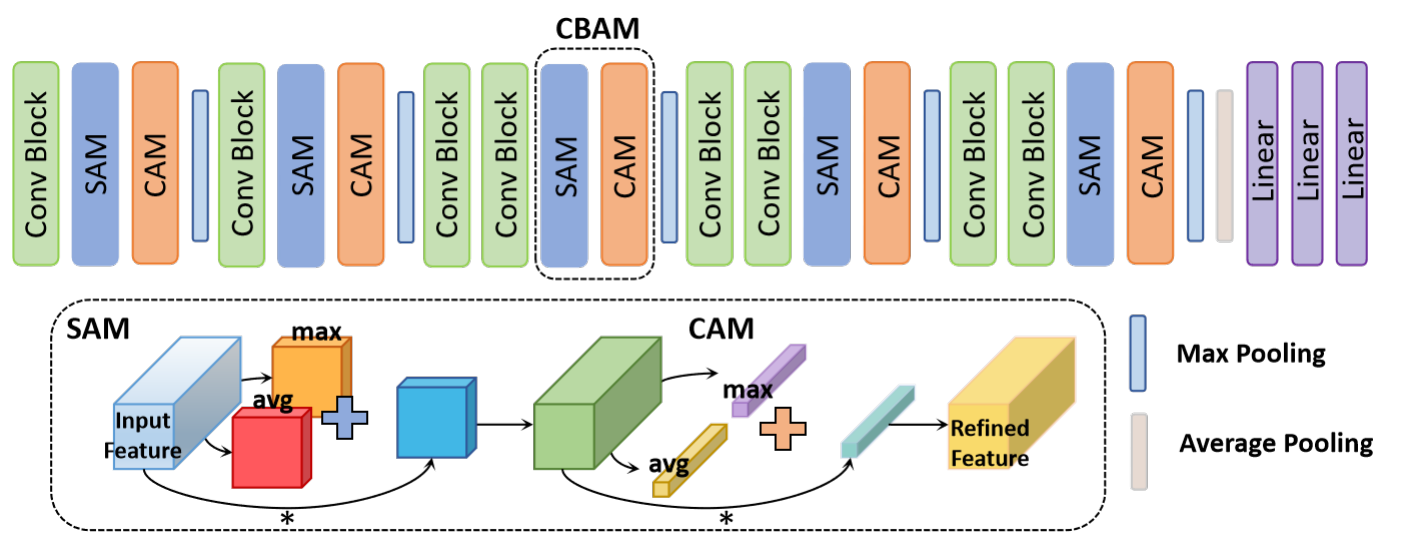}  
	\caption{The overall architecture of the methodology, which is the VGG equipped with CBAM module.}   
	\label{FIG:1}
 
\end{figure*}
\section{RELATED WORK}

\subsection{Event Representations}
Event data are usually converted into event-frame, time surface or grid-like representation for subsequent processing. The commonly employed event-frame methods adopted the constant time window or constant event number methods for converting events into image representations \cite{5,6,7}, thereby facilitating object recognition and orientation estimation through neural network. Traditional time surface usually stored the latest timestamp per-pixel \cite{delbruck2008frame}, which means the latest happening of activities possess more "intensity" in the image. However, simply using the individual time surface as representations exists problems such as motion blur and incomplete information expression. Different from that, Sironi \textit{et al}. proposed the histogram of averaged time surfaces \cite{sironi2018hats}, which utilizes the averaged time surfaces among the spatial-temporal domain as feature descriptor for object classification. Besides, Bai \textit{et al}. concatenated the latest temporal surface on two polarity channels with the event-count channel for comprehensively representing the outline and feature information of objects \cite{s1}. Deng \textit{et al}. incorporated multiple perspectives of the spatial-temporal event data and adopted the transformer architecture for feature fusion \cite{deng2021mvf}. And Zhu \textit{et al}. proposed a 4 channel image representation \cite{s2}, including the number and latest timestamps of positive and negative events per-pixel. Similarly, Alonso \textit{et al}. proposed a 6-channel image representation \cite{alonso2019ev}, which contains the mean and standard deviation value of the normalized timestamps of positive/negative events per-pixel together with the histogram of two polarity. Meanwhile, the event temporal image proposed by Fan \textit{et al}. mapped the events into image-like representations through activation function \cite{s3}, and they utilized an adaptive dense-to-sparse processing module to effectively balance detection efficiency and computational complexity across varying motion speed. Zhu \textit{et al}. proposed a discretized 3D volume representation assigning linear weighted to the surrounding events for the purpose of image accumulation \cite{zihao2018unsupervised}. Additionally, they introduced a joint loss function that integrates spatial smoothness loss with temporal unequal weighting loss.

\subsection{Network Architecture}

The commonly adopted networks are usually ResNet architecture \cite{he2016deep}, UNet \cite{ronneberger2015u}, MobileNetV2 \cite{sandler2018mobilenetv2}, MobileNetV3 \cite{howard2019searching}, InceptionV3 \cite{szegedy2016rethinking}, and they demonstrated the strong feature extraction capabilities. Besides, Deng \textit{et al}. proposed to comprehensively utilize different view fusion of the event data \cite{deng2021mvf}, such as the x-y, x-t, y-t dimensions, and processed the parallel data with a dual-structure of ResNet. Cannici \textit{et al}. proposed the Matrix Long Short-Term Memory (Matrix-LSTM) module for processing the event data \cite{cannici2020differentiable}, which is a differentiable and efficient end-to-end procedure. Baldwin \textit{et al}. innovatively proposed the "First-In-First-Out" buffer volume per-pixel for the event spike \cite{baldwin2022time}, whose stored logarithmic value of difference between timestamps is utilized for subsequent convolution and feature extraction. Chen \textit{et al}. proposed a hierarchical spatial relation module and the motion attention module \cite{chen2022ecsnet}, which enhance the capability of spatial and long-term temporal feature extraction of the event data respectively.

\section{EXPERIMENT DESIGN}
\subsection{Event Data}
SpikingJelly is an open-source deep learning based framework \cite{fang2023spikingjelly}, which possesses the superior extensibility, flexibility and user-friendliness. Especially, it provides numerous widely-used neuromorphic datasets, offers an intuitive user interface and integrated functions, which also greatly facilities users' opportunities for further adaptive development. Here, we utilize its embedded integrate approach for slicing the discrete events into frames \cite{fang2021incorporating}.

In this way, we acquire the two-channels frame representation through dividing the number of events into sequences of equal sizes. Among them, we assume the total number of event points as N, the slices of frames as T, and the floor operation $\lfloor$N/T$\rfloor$ represents the number of events in each slice. We utilize $n_L= \lfloor N/T \rfloor \cdot n $ and $n_R= \lfloor N/T \rfloor \cdot (n+1) $ as the left and right boundaries of the either selected slice, and the function $\Gamma_{p, x, y} (p_i, x_i, y_i)$ equals to 1 only when $(p, x, y) = (p_i, x_i, y_i)$. Therefore its formula can be expressed as the following (1).
\begin{equation}
\begin{aligned}
\mathcal Frame(n, p, x, y)=\sum_{i=n_L}^{n_R}\Gamma_{p, x, y} (p_i, x_i, y_i) 
\end{aligned}
\end{equation}

\subsection{Network Architecture}
Different from the architecture utilized in previous methods, we aim to propose a light-weight spatial-temporal learning framework for the event-based recognition. Fig.1 shows the overall network architecture, which is a VGG network equipped with the CBAM module \cite{simonyan2014very}. Among them, the Conv Block contains the convolution layer, batch normalization layer, and relu activation layer twice. And the CBAM module is composed of the SAM and CAM module sequentially, whose overall purpose is to refine the input feature so as to pay more attention on the specific and significant channels and spatial location.
\begin{table*}[h]

    \caption{Performance comparison of our method with other approaches. The Top-1 accuracy is highlighted in bold, and the second-best accuracy is underlined.}
    \centering
    \renewcommand{\arraystretch}{1.3}
    
    \begin{tabular}{ccccccccc}
        \toprule
        
		  \textbf{ImageNet}	&   \textbf{Method}	& \textbf{Event Representation} &\textbf{Data Augmentation}	& \textbf{CIFAR10-DVS}  & \textbf{N-Caltech101}    & \textbf{Classifier}  
       \\
			\midrule
  \multirow{6}{*}{Pretrained} & EST\cite{p1} & Grid-based  &\XSolidBrush	& $74.9$\%	& $83.7$\%		& ResNet-34	\\ 
~ &E2VID\cite{rebecq2019events} &Grid-based  &\XSolidBrush	& $51.5$\%	&\underline{$86.6$\%} 			&  ResNet-18	\\ ~ & M-LSTM\cite{cannici2020differentiable} & Ordered event sequences  &\CheckmarkBold	& $73.0$\%	& $85.7$\%			&  ResNet-34	\\~ & AMAE\cite{deng2020amae} & Event-based frame  &\CheckmarkBold	& $75.3$\%	& $85.1$\%			&  ResNet-34	\\ 
                    
                      ~      &     MVF-Net \cite{deng2021mvf}  & Event-based frame   &\XSolidBrush       & \underline{$76.2$\%}     & \textbf{87.1\%}	     & ResNet-34 \\

                   \cmidrule(lr){2-7}
                 ~ &  \textbf{Ours}    & Event-based frame	&\XSolidBrush  & \textbf{76.4\%}		& $85.0$\%		     
                             & VGG-CBAM  \\
                             \midrule
  \multirow{6}{*}{Not Pretrained} &   EST\cite{p1}	& Grid-based &\XSolidBrush  & \underline{$63.4$\%}		& \underline{$68.0$\%}					&  ResNet-34\\
  ~ &  E2VID\cite{rebecq2019events} &Grid-based  &\XSolidBrush	& $40.1$\%	& $57.6$\%			&  ResNet-18	\\
  ~ &   M-LSTM\cite{cannici2020differentiable} &Ordered event sequences &\CheckmarkBold	& $63.1$\%	& $67.2$\%		&  ResNet-34	\\
 ~ &  AMAE\cite{deng2020amae} &Event-based frame   &\CheckmarkBold	& $62.0$\%	& $66.8$\%			&  ResNet-34	\\
			  	              ~ &    MVF-Net\cite{deng2021mvf} &Event-based frame	 &\XSolidBrush	& $59.9$\%	& $66.3$\%				&  ResNet-34\\
                   \cmidrule(lr){2-7}
~ &\textbf{Ours}   & Event-based frame	&\XSolidBrush  & \textbf{71.3\%}		& \textbf{72.4}\%	     
                             & VGG-CBAM   \\

			\bottomrule


    \end{tabular}

    \label{tab: dataset}

\end{table*}
\subsection{Implementation Details}

On the CIFAR10-DVS dataset\cite{li2017cifar10}, we choose to train the model for 100 epochs by utilizing the cross-entropy loss for optimization. Among them, we utilize the Adam optimizer, and the learning rate value is fixed at 1E-4. During the training process, we set the batch size of the dataset at 128, and the slices of frames N at 20. While on the N-Caltech101 dataset\cite{orchard2015converting}, we utilize the Adam optimizer, and the learning rate value is fixed at 1E-5. Besides, the batch size is set as 128, together with 30 training epochs.

\section{RESULTS}

We evaluated our lightweight framework on the publicly available datasets: CIFAR10-DVS \cite{li2017cifar10} and N-Caltech101 \cite{orchard2015converting}. CIFAR10-DVS \cite{li2017cifar10} is collected through utilizing the dynamic vision sensor to do the repeated closed-loop smooth movement in front of the original CIFAR10 dataset. It contains ten different classes: airplane, automobile, bird, cat, deer, dog, frog, horse, ship and truck, and possesses 128×128 resolution together with average 1200ms duration time. Similarly, N-Caltech101 \cite{orchard2015converting} is collected by a rotary asynchronous time-based image sensor (ATIS), possessing 8246 samples and 101 classes. 


Table 1 demonstrates our framework's performance against five state-of-the-art ResNet-based methods, categorized by pretrained and non-pretrained approaches. Our VGG-CBAM architecture without pretraining achieves 71.3\% accuracy on CIFAR10-DVS and 72.4\% on N-Caltech101, while the pretrained model achieves 76.4\% accuracy on CIFAR10-DVS and 85.0\% on N-Caltech101. Compared to the residual structure of ResNet, VGG has a simpler and flexible structure for using the repeated stacking of small convolution cores, which has advantages like plug and play components. In this way, it can not only increase the nonlinear mapping but also improve the fitting expression ability of network. Simultaneously, because we adopt to integrate two-channels representation of the event data. Thus, compared to the original VGG module, the number of parameters in our framework has been favorably reduced 2.3\%, and the floating-point operations (FLOPs) has been averagely reduced 15.9 MFLOPs. 

\section{CONCLUSION}

In this work, we propose a novel spatial-temporal learning framework for the event-based object recognition and test its performance on two publicly available datasets. It leverages the SpikingJelly platform for the pre-processing of the event data, and utilizes the VGG equipped with the CBAM module as the classifier, which dynamically recalibrates feature responses through both channel and spatial attention mechanisms. 

The framework makes several key contributions to the field of event-based vision. First, we demonstrate a comparatively flexible and simple network architecture that achieves competitive accuracy on CIFAR10-DVS and N-Caltech101 datasets among the existing approaches.    Second, our comprehensive evaluation on standard event-based datasets shows robust performance across diverse object categories. Finally, our successful integration of attention mechanisms in this spatiotemporal pipeline opens new possibilities for efficient event-based recognition.

In the future, we expect to further reduce the parameters and computation efficiency of the framework,  responding to the growing need of neuromorphic sensors in autonomous drones and high-speed robotics. In these applications, where rapid motion detection and low-latency response are essential. Our approach provides a potential pathway for deploying event-based recognition systems in resource-constrained environments, demonstrating particular promise for applications requiring both high-speed vision processing and energy efficiency.  

\section*{Acknowledgment}

This work was supported in part by the Shenzhen Science and Technology Program JCYJ20230807114907015, and in part by the Research Grants Council of the Hong Kong Special Administrative Region, China, under Projects CityU 11214020 and CUHK-R4022-18.

\ifCLASSOPTIONcaptionsoff
  \newpage
\fi

\bibliographystyle{IEEEtran}
\bibliography{ref}

\begin{thebibliography}{10}
\providecommand{\url}[1]{#1}
\csname url@samestyle\endcsname
\providecommand{\newblock}{\relax}
\providecommand{\bibinfo}[2]{#2}
\providecommand{\BIBentrySTDinterwordspacing}{\spaceskip=0pt\relax}
\providecommand{\BIBentryALTinterwordstretchfactor}{4}
\providecommand{\BIBentryALTinterwordspacing}{\spaceskip=\fontdimen2\font plus
\BIBentryALTinterwordstretchfactor\fontdimen3\font minus \fontdimen4\font\relax}
\providecommand{\BIBforeignlanguage}[2]{{%
\expandafter\ifx\csname l@#1\endcsname\relax
\typeout{** WARNING: IEEEtran.bst: No hyphenation pattern has been}%
\typeout{** loaded for the language `#1'. Using the pattern for}%
\typeout{** the default language instead.}%
\else
\language=\csname l@#1\endcsname
\fi
#2}}
\providecommand{\BIBdecl}{\relax}
\BIBdecl

\bibitem{1}
T.~Serrano-Gotarredona and B.~Linares-Barranco, ``A 128 $\times$ 128 1.5\% contrast sensitivity 0.9\% fpn 3 $\mu$s latency 4 mw asynchronous frame-free dynamic vision sensor using transimpedance preamplifiers,'' \emph{IEEE Journal of Solid-State Circuits}, vol.~48, no.~3, pp. 827--838, 2013.

\bibitem{2}
C.~Brandli, R.~Berner, M.~Yang, S.-C. Liu, and T.~Delbruck, ``A 240$\times$ 180 130 db 3 $\mu$s latency global shutter spatiotemporal vision sensor,'' \emph{IEEE Journal of Solid-State Circuits}, vol.~49, no.~10, pp. 2333--2341, 2014.

\bibitem{3}
G.~Gallego, T.~Delbr{\"u}ck, G.~Orchard, C.~Bartolozzi, B.~Taba, A.~Censi, S.~Leutenegger, A.~J. Davison, J.~Conradt, K.~Daniilidis \emph{et~al.}, ``Event-based vision: A survey,'' \emph{IEEE transactions on pattern analysis and machine intelligence}, vol.~44, no.~1, pp. 154--180, 2020.

\bibitem{ercan2023evreal}
B.~Ercan, O.~Eker, A.~Erdem, and E.~Erdem, ``Evreal: Towards a comprehensive benchmark and analysis suite for event-based video reconstruction,'' in \emph{Proceedings of the IEEE/CVF Conference on Computer Vision and Pattern Recognition}, 2023, pp. 3943--3952.

\bibitem{shariff2024event}
W.~Shariff, M.~S. Dilmaghani, P.~Kielty, M.~Moustafa, J.~Lemley, and P.~Corcoran, ``Event cameras in automotive sensing: A review,'' \emph{IEEE Access}, 2024.

\bibitem{5}
R.~Ghosh, A.~Mishra, G.~Orchard, and N.~V. Thakor, ``Real-time object recognition and orientation estimation using an event-based camera and cnn,'' in \emph{2014 IEEE Biomedical Circuits and Systems Conference (BioCAS) Proceedings}.\hskip 1em plus 0.5em minus 0.4em\relax IEEE, 2014, pp. 544--547.

\bibitem{6}
M.~Iacono, S.~Weber, A.~Glover, and C.~Bartolozzi, ``Towards event-driven object detection with off-the-shelf deep learning. in 2018 ieee,'' in \emph{RSJ International Conference on Intelligent Robots and Systems (IROS)}, pp. 1--9.

\bibitem{7}
H.~Rebecq, R.~Ranftl, V.~Koltun, and D.~Scaramuzza, ``Events-to-video: Bringing modern computer vision to event cameras,'' in \emph{Proceedings of the IEEE/CVF Conference on Computer Vision and Pattern Recognition}, 2019, pp. 3857--3866.

\bibitem{delbruck2008frame}
T.~Delbruck \emph{et~al.}, ``Frame-free dynamic digital vision,'' in \emph{Proceedings of Intl. Symp. on Secure-Life Electronics, Advanced Electronics for Quality Life and Society}, vol.~1.\hskip 1em plus 0.5em minus 0.4em\relax Citeseer, 2008, pp. 21--26.

\bibitem{el2023cstr}
Z.~A. El~Shair, A.~Hassani, and S.~A. Rawashdeh, ``Cstr: a compact spatio-temporal representation for event-based vision,'' \emph{IEEE Access}, vol.~11, pp. 102\,899--102\,916, 2023.

\bibitem{p2}
Y.~Bi, A.~Chadha, A.~Abbas, E.~Bourtsoulatze, and Y.~Andreopoulos, ``Graph-based spatio-temporal feature learning for neuromorphic vision sensing,'' \emph{IEEE Transactions on Image Processing}, vol.~29, pp. 9084--9098, 2020.

\bibitem{p1}
D.~Gehrig, A.~Loquercio, K.~G. Derpanis, and D.~Scaramuzza, ``End-to-end learning of representations for asynchronous event-based data,'' in \emph{Proceedings of the IEEE/CVF International Conference on Computer Vision}, 2019, pp. 5633--5643.

\bibitem{simonyan2014very}
K.~Simonyan and A.~Zisserman, ``Very deep convolutional networks for large-scale image recognition,'' \emph{arXiv preprint arXiv:1409.1556}, 2014.

\bibitem{woo2018cbam}
S.~Woo, J.~Park, J.-Y. Lee, and I.~S. Kweon, ``Cbam: Convolutional block attention module,'' in \emph{Proceedings of the European conference on computer vision (ECCV)}, 2018, pp. 3--19.

\bibitem{fang2023spikingjelly}
W.~Fang, Y.~Chen, J.~Ding, Z.~Yu, T.~Masquelier, D.~Chen, L.~Huang, H.~Zhou, G.~Li, and Y.~Tian, ``Spikingjelly: An open-source machine learning infrastructure platform for spike-based intelligence,'' \emph{Science Advances}, vol.~9, no.~40, p. eadi1480, 2023.

\bibitem{li2017cifar10}
H.~Li, H.~Liu, X.~Ji, G.~Li, and L.~Shi, ``Cifar10-dvs: an event-stream dataset for object classification,'' \emph{Frontiers in neuroscience}, vol.~11, p. 309, 2017.

\bibitem{orchard2015converting}
G.~Orchard, A.~Jayawant, G.~K. Cohen, and N.~Thakor, ``Converting static image datasets to spiking neuromorphic datasets using saccades,'' \emph{Frontiers in neuroscience}, vol.~9, p. 437, 2015.

\bibitem{sironi2018hats}
A.~Sironi, M.~Brambilla, N.~Bourdis, X.~Lagorce, and R.~Benosman, ``Hats: Histograms of averaged time surfaces for robust event-based object classification,'' in \emph{Proceedings of the IEEE conference on computer vision and pattern recognition}, 2018, pp. 1731--1740.

\bibitem{s1}
W.~Bai, Y.~Chen, R.~Feng, and Y.~Zheng, ``Accurate and efficient frame-based event representation for aer object recognition,'' in \emph{2022 International Joint Conference on Neural Networks (IJCNN)}.\hskip 1em plus 0.5em minus 0.4em\relax IEEE, 2022, pp. 1--6.

\bibitem{deng2021mvf}
Y.~Deng, H.~Chen, and Y.~Li, ``Mvf-net: A multi-view fusion network for event-based object classification,'' \emph{IEEE Transactions on Circuits and Systems for Video Technology}, vol.~32, no.~12, pp. 8275--8284, 2021.

\bibitem{s2}
A.~Z. Zhu, L.~Yuan, K.~Chaney, and K.~Daniilidis, ``Ev-flownet: Self-supervised optical flow estimation for event-based cameras,'' \emph{arXiv preprint arXiv:1802.06898}, 2018.

\bibitem{alonso2019ev}
I.~Alonso and A.~C. Murillo, ``Ev-segnet: Semantic segmentation for event-based cameras,'' in \emph{Proceedings of the IEEE/CVF Conference on Computer Vision and Pattern Recognition Workshops}, 2019, pp. 0--0.

\bibitem{s3}
L.~Fan, Y.~Li, H.~Shen, J.~Li, and D.~Hu, ``From dense to sparse: Low-latency and speed-robust event-based object detection,'' \emph{IEEE Transactions on Intelligent Vehicles}, 2024.

\bibitem{zihao2018unsupervised}
A.~Zihao~Zhu, L.~Yuan, K.~Chaney, and K.~Daniilidis, ``Unsupervised event-based optical flow using motion compensation,'' in \emph{Proceedings of the European Conference on Computer Vision (ECCV) Workshops}, 2018, pp. 0--0.

\bibitem{he2016deep}
K.~He, X.~Zhang, S.~Ren, and J.~Sun, ``Deep residual learning for image recognition,'' in \emph{Proceedings of the IEEE conference on computer vision and pattern recognition}, 2016, pp. 770--778.

\bibitem{ronneberger2015u}
O.~Ronneberger, P.~Fischer, and T.~Brox, ``U-net: Convolutional networks for biomedical image segmentation,'' in \emph{Medical image computing and computer-assisted intervention--MICCAI 2015: 18th international conference, Munich, Germany, October 5-9, 2015, proceedings, part III 18}.\hskip 1em plus 0.5em minus 0.4em\relax Springer, 2015, pp. 234--241.

\bibitem{sandler2018mobilenetv2}
M.~Sandler, A.~Howard, M.~Zhu, A.~Zhmoginov, and L.-C. Chen, ``Mobilenetv2: Inverted residuals and linear bottlenecks,'' in \emph{Proceedings of the IEEE conference on computer vision and pattern recognition}, 2018, pp. 4510--4520.

\bibitem{howard2019searching}
A.~Howard, M.~Sandler, G.~Chu, L.-C. Chen, B.~Chen, M.~Tan, W.~Wang, Y.~Zhu, R.~Pang, V.~Vasudevan \emph{et~al.}, ``Searching for mobilenetv3,'' in \emph{Proceedings of the IEEE/CVF international conference on computer vision}, 2019, pp. 1314--1324.

\bibitem{szegedy2016rethinking}
C.~Szegedy, V.~Vanhoucke, S.~Ioffe, J.~Shlens, and Z.~Wojna, ``Rethinking the inception architecture for computer vision,'' in \emph{Proceedings of the IEEE conference on computer vision and pattern recognition}, 2016, pp. 2818--2826.

\bibitem{cannici2020differentiable}
M.~Cannici, M.~Ciccone, A.~Romanoni, and M.~Matteucci, ``A differentiable recurrent surface for asynchronous event-based data,'' in \emph{Computer Vision--ECCV 2020: 16th European Conference, Glasgow, UK, August 23--28, 2020, Proceedings, Part XX 16}.\hskip 1em plus 0.5em minus 0.4em\relax Springer, 2020, pp. 136--152.

\bibitem{baldwin2022time}
R.~W. Baldwin, R.~Liu, M.~Almatrafi, V.~Asari, and K.~Hirakawa, ``Time-ordered recent event (tore) volumes for event cameras,'' \emph{IEEE Transactions on Pattern Analysis and Machine Intelligence}, vol.~45, no.~2, pp. 2519--2532, 2022.

\bibitem{chen2022ecsnet}
Z.~Chen, J.~Wu, J.~Hou, L.~Li, W.~Dong, and G.~Shi, ``Ecsnet: Spatio-temporal feature learning for event camera,'' \emph{IEEE Transactions on Circuits and Systems for Video Technology}, vol.~33, no.~2, pp. 701--712, 2022.

\bibitem{fang2021incorporating}
W.~Fang, Z.~Yu, Y.~Chen, T.~Masquelier, T.~Huang, and Y.~Tian, ``Incorporating learnable membrane time constant to enhance learning of spiking neural networks,'' in \emph{Proceedings of the IEEE/CVF international conference on computer vision}, 2021, pp. 2661--2671.

\bibitem{rebecq2019events}
H.~Rebecq, R.~Ranftl, V.~Koltun, and D.~Scaramuzza, ``Events-to-video: Bringing modern computer vision to event cameras,'' in \emph{Proceedings of the IEEE/CVF Conference on Computer Vision and Pattern Recognition}, 2019, pp. 3857--3866.

\bibitem{deng2020amae}
Y.~Deng, Y.~Li, and H.~Chen, ``Amae: Adaptive motion-agnostic encoder for event-based object classification,'' \emph{IEEE Robotics and Automation Letters}, vol.~5, no.~3, pp. 4596--4603, 2020.

\end{thebibliography}

\end{document}